\documentclass[journal]{IEEEtran}

\usepackage{graphicx}
\usepackage{amsmath}
\usepackage{amssymb}
\usepackage{xcolor}
\usepackage{booktabs}
\usepackage{adjustbox}
\usepackage{tabularx}
\newcolumntype{Y}{>{\centering\arraybackslash}X}
\usepackage{multirow}
\usepackage{xspace}
\usepackage[detect-all,per-mode=repeated-symbol]{siunitx}
\DeclareSIUnit\pixel{px}
\usepackage{microtype}
\usepackage{algorithm2e}
\usepackage{caption}
\usepackage{subfig}
\usepackage{eccvabbrv}

\usepackage[pagebackref,breaklinks,colorlinks]{hyperref}
\usepackage[capitalize]{cleveref}

\begin{document}
\title{Cross-sensor self-supervised training and alignment for remote sensing}

\author{Valerio Marsocci, Nicolas Audebert%
\thanks{Valerio Marsocci is with the ESA $\Phi$-lab, Frascati (IT), and with the Cedric Lab, CNAM, Paris (FR), e-mail: valerio.marsocci@esa.int}%
\thanks{Nicolas Audebert is with the LASTIG, IGN, Saint-Mandè (FR) and with the Cedric Lab, CNAM, Paris (FR), e-mail: nicolas.audebert@ign.fr.}%

}

\markboth{Preprint}%
{Shell \MakeLowercase{\textit{et al.}}: Bare Demo of IEEEtran.cls for IEEE Journals}

\maketitle

\begin{abstract}
Large-scale ``foundation models'' have gained traction as a way to leverage the vast amounts of unlabeled remote sensing data collected every day. However, due to the multiplicity of Earth Observation satellites, these models should learn ``sensor agnostic'' representations, that generalize across sensor characteristics with minimal fine-tuning. This is complicated by data availability, as low-resolution imagery, such as Sentinel-2 and Landsat-8 data, are available in large amounts, while very high-resolution aerial or satellite data is less common. {To better leverage multi-sensor data}, we introduce cross-sensor self-supervised training and alignment for remote sensing (X-STARS). We design a self-supervised training loss, the Multi-Sensor Alignment Dense loss (MSAD), to align representations across sensors, even with vastly different resolutions, {through a contrastive patch-wise mechanism}. Our X-STARS can be applied to train models from scratch, or to adapt large models pretrained on \eg low-resolution EO data to new high-resolution sensors, in a continual pretraining framework.
We collect and release MSC-France, a new multi-sensor dataset, on which we train our X-STARS models, then evaluated on seven downstream classification and segmentation tasks. We demonstrate that X-STARS outperforms the state-of-the-art with less data across various conditions of data availability and resolutions.
\end{abstract}


\IEEEpeerreviewmaketitle

\section{Introduction}
\label{sec:intro}

\IEEEPARstart{A}{s} computing and data are becoming more available, large-scale pretrained models are becoming more common for Earth Observation (EO) \cite{visionpaper, Mendieta_2023_ICCV}. Inspired by the ``foundation models'' that excel in natural language processing \cite{chang2023survey} and computer vision \cite{jaegle2021perceiver}, large deep models have been tailored to EO data to consider the distinct challenges and opportunities of remote sensing (RS) and geospatial data \cite{Reed_2023_ICCV}.
One specificity of EO is that models should be \emph{sensor aware} (or \emph{sensor agnostic}). Indeed, RS data is sourced from numerous airborne and spaceborne sensors with different spatial resolutions, spectral bands and camera calibrations~\cite{visionpaper}. The ability to seamlessly adapt and exploit data from various sensors is crucial for ensuring consistent and reliable results, with models that generalize to new acquisitions \cite{visionpaper}. It also alleviates resource constraints, as one should be able to efficiently fine-tune large pretrained models on a new sensor with minimal data, without retraining from scratch for every sensor.
For these reasons, continual pretraining, \ie adding pretraining stages to the optimization of generic models to integrate data, has gained a lot of traction for RS~\cite{Reed_2022_WACV, Mendieta_2023_ICCV}, {as shown in \cref{sec:relwc}}.
By carefully training the model on new observations, better representations can be learned compared to full retraining, saving both time and resources.

\begin{figure}[]
\small
\centering
\includegraphics[width=.95\linewidth]{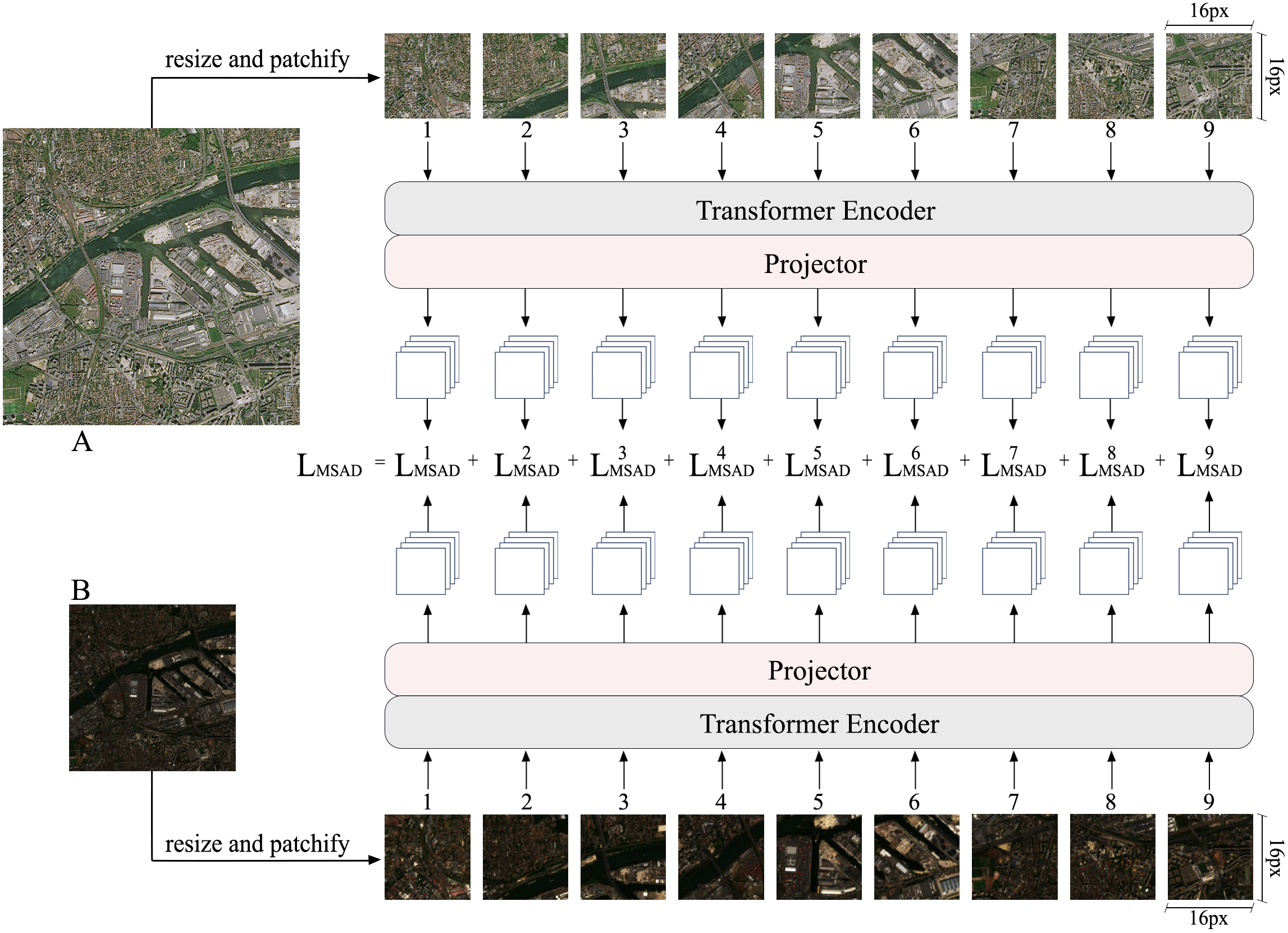}
\caption{Multi-sensor alignment dense (MSAD) loss aligns the features of the model applied on sensor $B$ to the features of the model applied on sensor $A$. The alignment is applied patchwise. The images have different sizes and are resized before the application of the model.}
\label{fig:msad}
\end{figure}

One way to achieve sensor agnosticism is with multimodality. While most multimodal models deal with a combination of texts and images \cite{radford2021learning}, EO offers many more possibilities:  multisource and multispectral imaging, Synthetic Aperture Radar (SAR), vector geodata, ground-level pictures, etc. To date, most works in multimodal learning for RS focused on combining active (SAR) and passive (optical) sensors \cite{hafner2022unsupervised}, underestimating the differences amongst optical sensors. Yet, optical images from different sensors can contain very different information due to changes in resolution, wavelengths and altitude. More specifically, they exhibit not only color variance, but also different aliasings, specularities, color dynamics, contrasts and blurs.
Therefore, models trained on one optical sensor tend to poorly generalize when applied to another one.
Furthermore, while a vast volume of Earth imagery is generated daily, access to open very high-resolution data remains limited. Most available datasets consist of mid to low-resolution images with a ground sampling distance (GSD) between 10 and \SI{60}{\meter}, such as Sentinel-2 (10-\SI{60}{\meter\per\pixel}) or Landsat-8 (\SI{30}{\meter\per\pixel}). Adaptation techniques could allow for training large-scale models on low-resolution datasets and then adapting them on-demand to scarcer high-resolution data {(see also \cref{sec:relwb})}.

To this end, we introduce an algorithm for \textit{Cross-sensor Self-supervised Training and Alignment of Remote Sensing data} (\emph{X-STARS}) {to effectively fine-tune large-scale models pretrained on an optical sensor to another one.}
X-STARS combines a contrastive self-supervised learning (SSL) with a novel sensor alignment objective: Multi-Sensor Alignment Dense loss (MSAD). The latter allows to capture the common semantics of local image patches independently from scale and sensors using knowledge distillation.
Masked image modeling (MIM), {both as continual pretraining \cite{Mendieta_2023_ICCV} and when pretraining from scratch ~\cite{cong2022satmae, Reed_2023_ICCV}}, requires large high-resolution datasets \cite{Mendieta_2023_ICCV}, X-STARS can deal with any image resolution. This allows us to improve on models that have been trained on low-resolution images only, using a fraction of the high-resolution data used for MIM.
{Unlike similar contrastive methods, X-STARS is properly shaped for EO multimodal tasks \cite{wang2021dense, iskender2023improvingdensecontrastivelearning}. and it is able to perform strongly on different downstream tasks \cite{chen2021self}. }
X-STARS can be applied both as an end-to-end self-supervised for from-scratch pretraining objective, or as \textit{a posteriori} fine-tuning loss for continual pretraining.

To train X-STARS, we collect Multi-Sensors Cities France (MSC-France), a new multimodal EO dataset that includes four sensors from low to very high-resolution imagery. %
We show that this approach outperforms existing pretrained models on downstream tasks across many different sensors and training setups (backbones, from-scratch training and continual pretraining). We establish a new state-of-the-art (SOTA) on various downstream tasks of classification and semantic segmentation of EO imagery.
In summary, our contributions are:
\begin{itemize}
  \setlength{\itemsep}{0pt}%
  \setlength{\parskip}{0pt}%
 \item MSC-France, a novel dataset of $\approx$5000 image triplets covering most French cities with three different sensors (SPOT-6, Landsat-8, Sentinel-2), and a subset with several high-resolution pairs (SPOT-6, BDORTHO);
 \item Multi-Sensor Alignment Dense loss (MSAD), a cross-sensor contrastive loss that aligns representations learned from different sensors, even with different GSDs;
  \item We train self-supervised models and improve performance on several downstream EO benchmarks with 10 to 100$\times$ fewer data compared to previous works and use continual pretraining to adapt pretrained models to new sensors.%
\end{itemize}

\section{Related Works}
\label{sec:relworks}  

\subsection{Remote Sensing Self-Supervised Learning}

RS is a fertile playground for SSL, as large amounts of (unlabeled) EO data are available worldwide. Many large-scale datasets have been published in the last few years, some labeled such as Dynamic EarthNet~\cite{toker2022dynamicearthnet} for change detection, BigEarthNet for scene classification~\cite{sumbul2019bigearthnet}, OpenEarthMap for land cover mapping~\cite{xia2023openearthmap}, and many others unlabeled~\cite{manas2021seasonal,cong2022satmae,christie2018functional}.
SSL approaches from traditional computer vision have been adapted to address the specifics of EO~\cite{wang2022self, tao2022self}. \cite{marsocci2021mare} adapts Online Bag-of-Words~\cite{gidaris2020online} to train without labels the backbone of a segmentation network. Contrastive learning is a staple of SSL for RS: SauMoCo~\cite{kang2020deep} applies contrastive learning on spatially augmented views of an image; \cite{stojnic2021self} train EO models with contrastive multiview coding  \cite{tian2019contrastive} on three large datasets with both RGB and multispectral bands;
\cite{zhang2022false} tunes the negative sampling strategy of contrastive learning to take into account the intra-class diversity of RS images; \cite{saha2022unsupervised} adapts contrastive learning for semantic segmentation of EO images; %
Seasonal Contrast (SeCo)~\cite{manas2021seasonal} uses contrastive learning by treating two images of the same area at different times as different views; CACo \cite{Mall_2023_CVPR} extends this principle by incorporating more levels of seasonal differences. Other works use pretext tasks, such as inpainting~\cite{tao2020remote} or reconstructing visible wavelengths from the other bands~\cite{vincenzi2021color}.
MIM is also popular: SatMAE \cite{cong2022satmae} trains a masked autoencoder (MAE) \cite{he2022masked}, properly adapted to deal with multi-temporal data. Scale-MAE \cite{Reed_2023_ICCV} deals with images of various resolutions using band filter decoding and scale-equivariant positional encoding.

While these approaches are effective in learning representations on unlabeled data, practical use remains limited to their poor generalization to sensors outside the training set. To this end, continual pretraining is a promising avenue. In particular, GFM~\cite{Mendieta_2023_ICCV} built upon the continual pretraining strategy from \cite{Reed_2022_WACV} to train geospatial ``foundation models''.

\subsection{Multimodal learning}
\label{sec:relwb}

Multimodal learning, especially image and text, has been boosted by the release of large vision-language models such as CLIP \cite{radford2021learning}. The main challenge of multimodal learning is to learn \emph{aligned} representations of the two modalities that preserve their common information, without discarding the modality-specific knowledge. In the following years, different works tried to tackle the limits with different strategies, such as softening the objective loss \cite{gao2023softclip, gao2022pyramidclip}, alternative losses \cite{zhai2023sigmoid} or combining the similarity loss with masking \cite{yang2023attentive, dong2023maskclip}. RS has not escaped this trend~\cite{liu2023remoteclip, wen2023vision, Singha_2023_CVPR, hu2023rsgpt}. 
In RS, multimodality takes its root from data fusion and therefore is generally envisioned as mixing SAR and optical data \cite{chen2021self, hafner2022unsupervised, li2022deep} or optical data and vector geodata \cite{audebert_joint_2017}. Recently, some works tried to use other combinations of data. For example, in \cite{deuser2023sample4geo}, the authors propose a contrastive framework to compare street views and satellite images. However, even two different optical sensors can be considered as different modalities since their spectral and spatial characteristics can vary wildly. For example, \cite{swope2021representation} collects a high-resolution dataset from three different sensors (NAIP, Pleiades, SPOT-6) and argues for training a model with a combination of sensors for better generalization. In the same idea, \cite{machado2020airound} and \cite{garioud2023flair} introduce datasets in which satellite images are enhanced with multi-view images, from different sources (\eg airborne or street view).
Yet, as multimodal data might not be available at training time, we design our approach so that it can be applied both to train from scratch or as a post-hoc adaption to fit an existing model on a new sensor.

\subsection{Continual pretraining and knowledge distillation}
\label{sec:relwc}

Knowledge distillation is a popular framework to train models by using representations from a strong ``teacher'' model as learning targets for a weaker ``student'' model~\cite{gou2021knowledge}. It has been used to reduce the size of a model by using a student architecture smaller than the teacher~\cite{hinton_distilling_2015}, to improve classification performance by softening the hard targets and extracting the ``dark knowledge'' from the teacher logits~\cite{xu2020knowledge} or self-train models by aligning features from multiple views of the same observation~\cite{caron2021emerging, oquab2023dinov2, chen2023sssd}. In our work, we use knowledge distillation as a cross-modal alignment in the spirit of~\cite{thoker_cross-modal_2019}, \ie as a way {to} align multiple views coming from \emph{different sensors}, instead of different augmentations.

Using knowledge distillation to refine an existing model without supervision is akin to continual pretraining. Continual pretraining refers to the practice of adding new pretraining stages to an existing model, that aim to improve and/or specialize its representations on task-specific data. This was first introduced to fine-tune the abilities of large language models on specialized domains \cite{gururangan2020don, liu2021continual, han2020econet} and to fine-tune self-supervised vision models on medical imagery~\cite{kalapos2022self}. More broadly, continual pretraining has been found to improve the performance of self-supervised models on many tasks~\cite{Reed_2022_WACV}. 
As RS data is available as large but unlabeled datasets, this has become a popular technique to specialize large-scale models for remote EO~\cite{mendieta2023gfm}. \cite{marsocci2023continual,moieez2023} \eg mix SSL and continual learning to integrate new observations, ending up in a continual pretraining framework to deal with non-stationary RS datasets.
In this work, we show that knowledge distillation can be an effective tool for multimodal alignment as a continual pretraining objective.

\begin{figure*}[t]
\small
\centering
\subfloat[\centering Pretraining from scratch]{{\includegraphics[width=8cm]{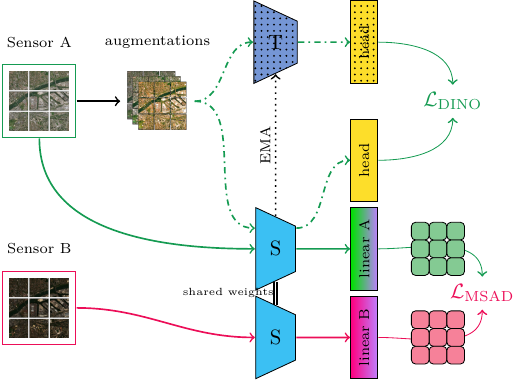} }}%
\qquad
\subfloat[\centering Continual pretraining]{{\includegraphics[width=8cm]{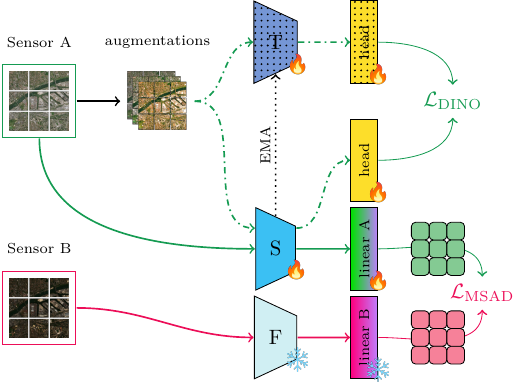} }}%
\caption{Training strategies using X-STARS. On the left (a), pretraining from scratch adds the Multi-Sensor Alignment Dense loss to the standard DINO self-supervised training scheme. The teacher is updated by an exponential moving average (EMA) of the student. On the right (b), the continual pretraining approach uses a frozen backbone $F$ to extract the features on which to apply the MSAD loss. The DINO student and teacher are initialized to the same weights as $F$.}
\label{fig:model}
\end{figure*}

\section{SSL and cross-sensor alignment}
\label{sec:method}

Our goal is to design a self-supervised training scheme that can deal with multiple sensors, even with different resolutions. To do so, we divide the loss function in two:
\begin{itemize}
    \item an off-the-shelf SSL objective, \eg DINO~\cite{caron2021emerging},
    \item our novel scale-aware alignment loss (MSAD).
\end{itemize}

Given a multimodal dataset $\mathcal{D}$ and a Vision Transformer (ViT) parametrized by its weights $\theta$, we minimize:
\begin{equation}
    \mathcal{L}(\mathcal{D}; \theta) = L_\text{SSL}(\mathcal{D}; \theta) + \lambda \cdot L_\text{MSAD}(\mathcal{D}; \theta)
    \label{eq:xstars}
\end{equation}

where $L_\text{SSL}$ is a contrastive self-supervised loss, $L_\text{MSAD}$ is our Multi-Sensor Alignment Dense loss and $\lambda \geq 0$ is a weighting hyperparameter.

We design the $\mathcal{L}_\text{MSAD}$ as a contrastive loss that learns \emph{invariant} representations across sensors, \ie representations that are sensor agnostic. {We took inspiration from dense contrastive learning from classical computer vision \cite{wang2021dense, iskender2023improvingdensecontrastivelearning}, to learn sensor-agnostic representations of EO images.} {Using such a loss, the contrastive framework can also work effectively on datasets of very diverse resolutions.} We collect a new multi-sensor dataset (MSC-France) of paired acquisitions, presented in Section \ref{sec:data}. {Unlike most of the methods, MSAD can be used on top of any SSL objective. We use DINO} in this work for its strong emerging representations~\cite{caron2021emerging}.
To tackle the scarcity of very high-resolution RS data, we will show that our framework can be used to train large models on low-resolution data, and then adapt them later on to fewer high-resolution acquisitions.

\subsection{Contrastive loss}
\label{sec:dino}
Most contrastive losses pass two different augmentations of the same image to a student network $S$ and a teacher network $T$ that share their architecture. More precisely, multiple views $x$ are generated, containing two global views and several local -- \eg cropped -- views at different resolutions. This multi-scale approach is pivotal for our approach, based on cross-sensor alignment.
All crops are passed through the student while only the global views are passed through the teacher, therefore encouraging ``local-to-global” correspondences.
The output of the teacher network is centered with a mean computed over the batch. Each network outputs a $K$-dimensional feature vector transformed into probability distributions  $P_s$ and $P_t$ by a softmax with a temperature $\tau$ over the feature dimension:

\begin{equation}
P_{\theta_s}(x)^{(i)}=\frac{\exp \left(g_{\theta}(x)^{(i)} / \tau\right)}{\sum_{k=1}^{K} \exp \left(g_{\theta}(x)^{(k)} / \tau\right)},
\end{equation}
where $g_{\theta}$ is a network (\ie $S$ or $T$).
With a fixed teacher, their similarity is then measured with a cross-entropy loss:
\begin{equation}
\min_{\theta{s}} -P_{\theta_t}(x)\log[P_{\theta_s}(x)]
\end{equation}

Competitive SSL objectives, such as MIM, are ill-suited to our work as they require high-resolution images to be effectively trained on RS data \cite{Mendieta_2023_ICCV}. Indeed, low-resolution patches are too easy to reconstruct, making the model perform poorly on downstream tasks, as we will show in \cref{sec:comp}.

\subsection{Multi-Sensor Alignment Dense loss}
\label{msad}

Consider a dataset of acquisition pairs from two different sensors $A$ and $B$. We denote $X_A$ and $X_B$ images from these two sensors. Each image from one sensor (e.g., $X_A$) covers the exact same geographical area as the other sensor (e.g., $X_B$). We define $v^A = \psi_A(S(X_A))$ and $v^B = \psi_B(S(X_B))$ the representations  -- \ie global token -- obtained by passing the image through a Vision Transformer $S$ and a projection $\psi$. In practice, $\psi_A$ and $\psi_B$ are two linear layers. The backbone $S$ is trained using the contrastive InfoNCE loss so that $v_A$ and $v_B$ are invariant to their respective sensor:

\begin{equation}
    \mathcal{L}_{\text{InfoNCE}} = -\frac{1}{2 N} \sum_{i=1}^N \sum_{j=1}^N {y}_{i j}^A \cdot \log p_{i j}^A + {y}_{i j}^B \cdot \log p_{i j}^B
\end{equation}

where ${y}_{i j}^A$ and ${y}_{i j}^B$ are defined as Kronecker Deltas $\delta_{ij}$, and $p_{i j}^A$ and $p_{i j}^B$ are defined by:

\begin{equation}
    p_{i j}^A=\frac{\exp \left(\operatorname{sim}\left(v^A_i, v^B_j\right) / \tau\right)}{\sum_{k=1}^N \exp \left(\operatorname{sim} \left(v^A_i, v^B_k\right) / \tau\right)}
\end{equation}

and same for $p_{i j}^B$, with $\tau > 0$ a temperature parameter and $\operatorname{sim}$ the cosine similarity in feature space.
However, using the global token from a ViT as the embedding, the alignment operates only on global information and does not consider local semantics.
To integrate local information into the alignment, we instead use the patches from the last Transformer layer. Denoting $v[t]$ the $T$ tokens from $v^A$ and $v^B$, we sum the patchwise components of the loss on each token:

\begin{equation}
    p_{i j}^A=\sum_{t=1}^T \frac{\exp \left(\operatorname{sim}\left(v^A_i[t], v^B_j[t]\right) / \tau\right)}{\sum_{k=1}^N \exp \left(\operatorname{sim} \left(v^A_i[t], v^B_k[t]\right) / \tau\right)}
\end{equation}

where $T$ is the number of tokens. %
By averaging over local tokens, the contrastive loss now compares not only the global semantics but also the local ones. This is especially important for EO, as similar scenes can be comprised of very different objects or patterns due to high intra-class diversity~\cite{6827949, Marsocci_2023_CVPR}. 

A drawback of InfoNCE is that hard labeling only defines pairs as ``positive'' or ``negative''. Softening the targets has been shown to be useful in CLIP, \eg when text and image are not perfectly aligned~\cite{NEURIPS2022_e9882f7f, gao2023softclip}. This is especially important in EO, since acquisitions of the same area by different sensors are generally not simultaneous. Therefore, changes, either seasonal or structural, can happen. In addition, some unpaired acquisitions can look extremely similar, \eg large forests with the same type of trees but in different areas. Therefore, we use softened targets to relax the similarity constraints and allow that some unpaired tokens of different sensors might be similar and vice-versa.
In practice, the softened targets $\widetilde{\boldsymbol{y}}_i^l$ for the $i$-th pair can be formulated as:
\begin{equation}
    \widetilde{\boldsymbol{y}}_i^l=(1-\alpha) \boldsymbol{y}_i^l+\alpha /(N-1),
\end{equation}
where $\alpha$ is a fixed smoothing parameter.

Wrapping everything together, our final MSAD loss is:
\begin{equation}
    \mathcal{L}_\text{MSAD}=-\frac{1}{2 N} \sum_{i=1}^N \sum_{j=1}^N\left(\widetilde{y}_{i j}^A \cdot \log \left(p_{i j}Al\right)+\widetilde{y}_{i j}^B \cdot \log \left(p_{i j}^B\right)\right)
\end{equation}

\subsection{X-STARS}

\begin{algorithm}[t]
\renewcommand{\KwSty}[1]{\textnormal{\textcolor{green!30!black}{\bfseries #1}}\unskip}
\newcommand\commentfont[1]{\scriptsize\ttfamily\textcolor{blue!70!black}{#1}}
\SetCommentSty{commentfont}
\caption{Training of X-STARS from scratch.}\label{alg:cap}
\KwData{dataset $\mathcal{D} = \{(X_A, X_B)_{1\leq i\leq N}\}$}
\KwIn{$\lambda > 0$, ViT backbone $S$}
Initialize at random $S$ and projectors $\psi_A, \psi_B$\;
Set $T$ as an exponential moving average of $S$\;
\While{training}{
Sample a pair $X_A, X_B \in \mathcal{D}$\;
\tcc{DINO loss}
\For{$X$ in $X_A, X_B$}{%
    $X_1, X_2 \gets  \text{augment}(X)$ \tcp*[r]{aug. views}
    $f_{1}^{s}, f_{2}^{s} \gets S(X_1), S(X_2)$ \tcp*[r]{student feats}
    $f_{1}^{t}, f_{2}^{t} \gets T(X_1), T(X_2)$ \tcp*[r]{teacher feats}
    $L_{\text{D}} \gets \text{DINO}(f_{1}^{s}, f_{2}^{s}, f_{1}^{t}, f_{2}^{t})$\;
}%
\tcc{MSAD loss}
$s, l \gets \psi_S(S(X_A)), \psi_L(S(X_B))$ \tcp*[r]{projection}
$L_{\text{MSAD}} \gets \text{MSAD}(s, l)$\;
\tcc{Backpropagation}
SGD on $S$ for $L \gets L_{\text{D}} + \lambda L_{\text{MSAD}}$\;
}
\end{algorithm}

X-STARS can be used in two setups:
\par \textit{\textbf{i) pretraining from scratch}} is the standard SSL setup. The backbone $S$ is initialized and random and trained end-to-end using X-STARS by combining DINO and MSAD loss, as shown in Figure \ref{fig:model} (left) and detailed in Algorithm \ref{alg:cap}.
\par \textit{\textbf{ii) Continual pretraining }} improves an existing large pretrained model on a new task through a secondary pretraining stage.
In our case, we assume that a pretrained model -- the domain teacher $F$ -- is available, pretrained on domain $A$. We initialize the student $S$ and teacher $T$ used by DINO to the same architecture and weights as $F$. However, $F$ is frozen throughout the training process. We assume that, at least for a subset of images from domain $A$, there exists a paired dataset of images $(X_A, X_B)$ of images respectively from pretraining sensor $A$ and a new sensor $B$. We then train $S$ and $T$ as shown in Figure \ref{fig:model} (right). $S$ and $T$ receives the standard DINO loss on the images $X_B$ of the new sensor $B$. In addition, $F$ extracts the features $l$ from the corresponding images $X_A$ from the original sensor $A$. These images should have similar characteristics to those used to pretrain $F$. $S$ is then also optimized through the MSAD loss computed between the representations of the student $v^A = \psi_A(S(X_A))$ and the frozen domain teacher $v^B = \psi_B(F(X_B))$.

\section{MSC-France dataset}
\label{sec:data}

To pretrain our model, we collected a multi-sensor dataset of aerial and satellite imagery over France. There were only three multi-optical sensor datasets in the literature: Contrastive Sensor Fusion (CSF) \cite{swope2021representation}, AiRound \cite{machado2020airound} and FLAIR \cite{garioud2023flair}. However, CSF contains few images, and mostly small patches from high-resolution images (SPOT-6, NAIP, Pléiades). Meanwhile, AiRound combines Sentinel-2 images with airborne data with very different resolutions (\SI{0.3}{\meter}--\SI{4800}{\meter} GSD). Finally, FLAIR has a strong resolution gap between the two sensors (\SI{10}{\meter} GSD of Sentinel-2 vs \SI{0.25}{\meter} GSD of BDORTHO). These reasons make these datasets impractical for our needs.

\begin{figure}[]
\small
\centering
\includegraphics[width=1.\linewidth]{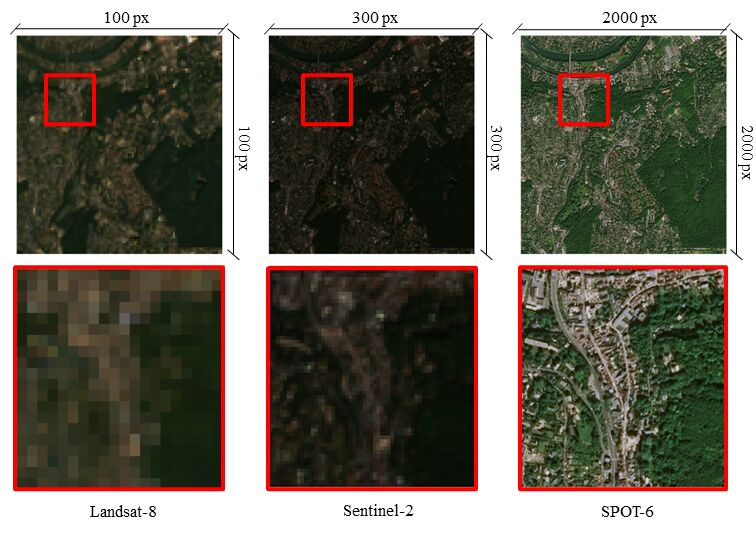}
\caption{A triplet from MSC-France dataset. On the first line the resized images are shown. On the second, we focus on a random area to show the different resolutions.}
\label{fig:msc2}
\end{figure}

\textbf{Multi-Sensor Cities France (MSC-France)} aggregates RGB images from three sensors: Sentinel-2 (\SI{10}{\meter\per\pixel}), Landsat-8 (\SI{30}{\meter\per\pixel}) and SPOT-6 (\SI{1.5}{\meter\per\pixel}\footnote{Pansharpened from the \SI{6}{\meter\per\pixel} multispectral image.}). 
It contains 4496 triplets of Sentinel-2, Landsat-8 and SPOT-6 images. An example is shown in Figure \ref{fig:msc2}. Specifically, Sentinel-2 images represent the level 2A reflectances values, processed by Sen2Cor. We considered only the second (blue), third (green) and fourth (red) bands, with a spatial resolution of \SI{10}{\meter\per\pixel}.
Landsat-8 images are captured by Landsat 8 OLI/TIRS sensors, represent the atmospherically corrected surface reflectance. As for Sentinel-2, we considered only the second (blue), third (green) and fourth (red) bands, with a spatial resolution of \SI{30}{\meter\per\pixel}.
SPOT-6 \textcopyright AIRBUS DS (2018) images are pansharpened from the \SI{6}{\meter\per\pixel} multispectral to the \SI{1.5}{\meter\per\pixel} panchromatic and orthorectified by IGN.
Landsat-8 and Sentinel-2 were downloaded from Google Earth Engine. We filtered the images selecting only the on with $<0.1\%$ cloud coverage, in the timeframe of summer of 2018. Then starting from the last available, we filled the areas covered with clouds with the precedent available info. 
As said, each triplet insists specifically on the same area, leading to different sizes among the different sensors:
\begin{itemize}
    \item $2000\times2000$ for SPOT-6 (GSD \SI{1.5}{\meter\per\pixel});
    \item $300\times300$ for Sentinel-2 (GSD \SI{10}{\meter\per\pixel});
    \item $100\times100$ for Landsat-8 (GSD \SI{30}{\meter\per\pixel}).
\end{itemize}
The dataset covers 12 major French cities and their suburbs (Paris, Nice, Toulouse, Bordeaux, Strasbourg, Lyon, Rennes, Lille, Marseille, Montpellier, Grenoble, and Nantes). This balances the urban-to-rural area ratio, while keeping a large diversity of backgrounds useful for downstream tasks. For example, the Toulouse region includes part of the Pyrenees mountains, while Nice also captures the seaside, and Paris imagery encompasses the city and its nearby forests. From MSC-France, we intentionally sample outside of over-represented but less distinctive areas, such as the sea and agricultural areas. %

\begin{figure}[h!]
\small
\centering
\includegraphics[width=1.\linewidth]{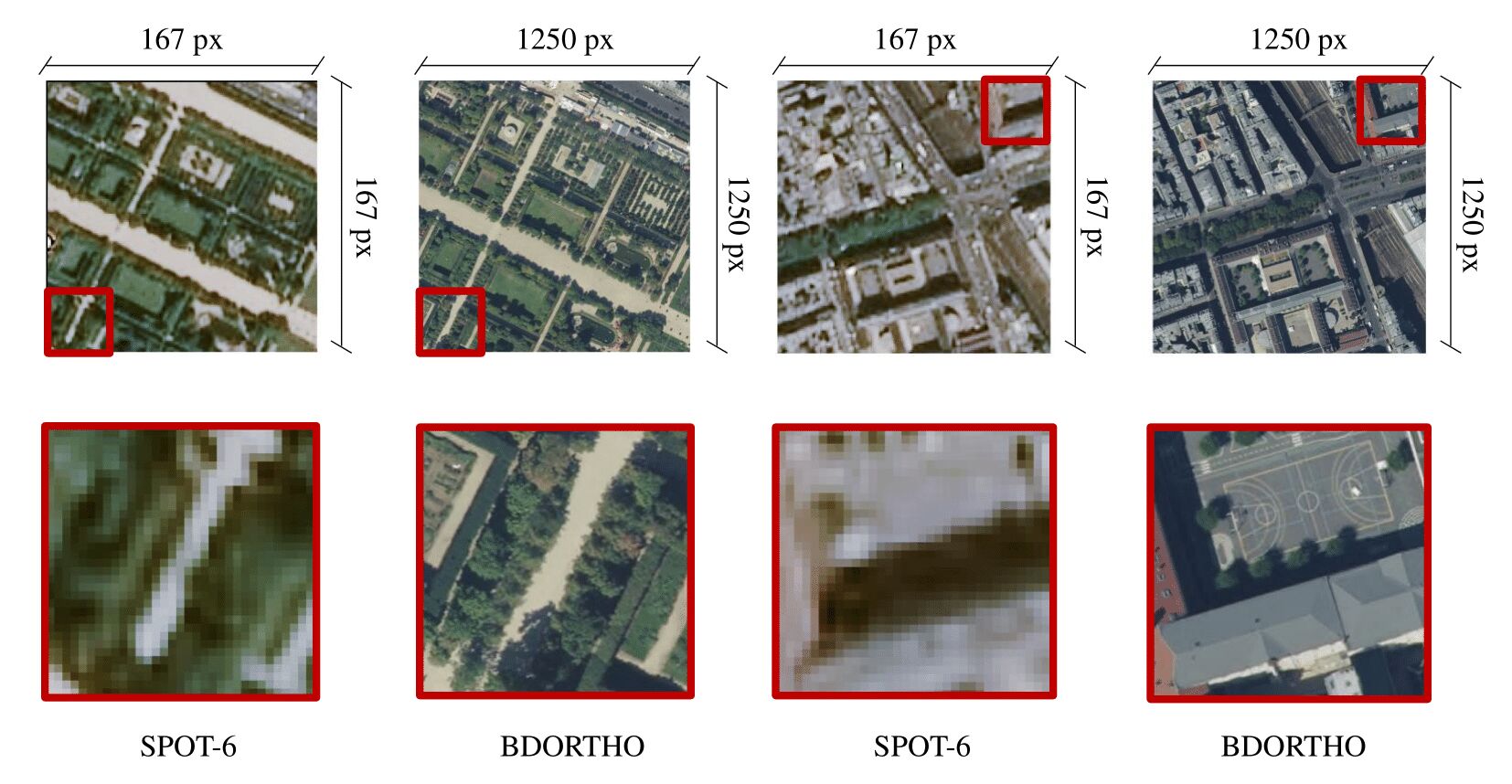}
\caption{Two pairs of the Multi-View Île-de-France. On the first line, the resized images are shown. On the second, we focus on a random area to show the different resolutions.}
\label{fig:bd2}
\end{figure}

We introduce also another dataset: \textbf{Multi-View Île-de-France}. It consists of 45,200 pairs of aerial BDORTHO\footnote{\url{https://geoservices.ign.fr/bdortho}}, with a GSD of \SI{0.2}{\meter\per\pixel}, and satellite SPOT-6, with a GSD of \SI{1.5}{\meter\per\pixel}. It covers all the Île-de-France, (\ie the 75th, the 92nd, the 93rd, the 94th, and the 95th departments). Each BDORTHO image is $1250\times1250$ and each SPOT-6 is $167\times167$.
In Figure \ref{fig:bd2}, two examples are presented.

\section{Experiments}
\label{sec:expres}

We compare X-STARS to SOTA SSL models for EO data: Scale-MAE \cite{Reed_2023_ICCV}, SatMAE \cite{cong2022satmae}, {CROMA \cite{fuller2024croma}, DINO-MC \cite{wanyan2023dinomc}, DINO-TP \cite{wanyan2023dinomc}} and SeCO \cite{manas2021seasonal}. We use the public weights for these models. {Because CROMA is trained on multi-spectral and SAR images, we consider only the ``optical encodings'' (\ie the weights of the multi-spectral encoder) and we zero-pad the channels not available in RGB datasets.} Moreover, we also report the results of {a vanilla DINO \cite{caron2021emerging}, }SatMAE and Scale-MAE trained\footnote{The hyperparameters are taken from the official GitHub repositories.} on our MSC-France dataset, to understand the impact of different data on the performance.

\subsection{Experimental setup}

For the ``pretraining'' setup of X-STARS, we train a ViT-L/16 and a ResNet50 backbone on 8 NVIDIA V100 GPUs for 800 epochs. We use only the DINO loss when one sensor is involved, or the full X-STARS loss with MSAD when two or more sensors are available.
For the ``continual pretraining'' setup, the model is first pretrained, and then adapted using the MSAD loss for 400 additional epochs. {We perform the default augmentations from the DINO paper for the DINO step and apply no augmentation for the MSAD step.}
We use a batch size of 32 images/GPU in all the experiments. All the input images are resized to $224\times224$. DINO hyperparameters are set as per \cite{caron2021emerging}. We use a label smoothing $\alpha = 0.3$ and a weight $\lambda = 0.1$ for MSAD, obtained through cross-validation. 

\begin{table*}[]
\caption{Metrics (top-1 accuracy and mIoU) on seven downstream classification and segmentation tasks. {For the dataset, }C denotes Classification tasks, S denotes Semantic Segmentation. {V denotes ViT, R denotes ResNet and S denotes Swin \cite{liu2021swin}.} Results are reported for k-NN (average over multiple $k$)/linear probing. $A \rightarrow B$ denotes models trained using continual learning, with an alignment from sensor $A$ to sensor $B$. %
}
\centering
\begin{adjustbox}{width=1\textwidth}
\label{tab:comp}
\setlength{\tabcolsep}{2pt}
\begin{tabular}{ccccccccccccc}
\toprule
& Model & Encoder & Pretrain Data & Init Weights & EuroSAT (C) & Worldstrat (C) & UC-Merced (C) & CV-BrCT (C) & FLAIR (S) & SN-8 (S) & CLC (S) \\
\midrule
\multirow{12}{*}{\rotatebox[origin=c]{90}{From scratch}} 
& Supervised & V & ImageNet & random & 89.3/88.7 & 70.1/69.2 & 90.4/91.4 & 74.2/74.6 & 34.8 & 63.8 & 14.1 \\
& SeCo \cite{manas2021seasonal} & R & SeCo dataset & random & 62.6/93.9 & 56.4/69.9 & 36.5/91.8 & 42.2/71.8 & 34.6 & 59.9 & 16.2 \\
& {CROMA} \cite{fuller2024croma} & V & SSL4EO-S12 & random & 87.3/85.2 & 63.1/65.1 & 71.1/73.4 & 65.2/63.6 & 36.2 & 61.1 & 15.1\\
& {DINO} & V & S2-LS & random & 93.6/94.2 & 69.3/70.1 & 89.5/90.5 & 73.8/74.4 & 46.3 &  65.4 & 18.5 \\
& {DINO-MC} \cite{wanyan2023dinomc} & V & SeCo dataset & random & 94.3/95.6 & 69.7/70.9 & 89.1/91.1 & 74.5/74.6 & 47.5 & 65.4 & 18.9 \\
& {DINO-TP} \cite{wanyan2023dinomc} & V & SeCo dataset & random & 93.2/94.9 & 69.3/69.9 & 91.2/91.8 & 74.8/74.5 & 47.3 & 65.8 & 18.5 \\
& SatMAE  \cite{cong2022satmae} & V & fMoW-Sentinel & random & 91.1/94.6 & 68.8/69.4 & 82.4/80.0 & 69.9/64.9 & 46.7 & 67.8 & 20.2 \\
& SatMAE & V & S2-LS & random & 83.4/88.5 & 66.1/71.4 & 65.4/75.0 & 68.4/65.4 & 35.1 & 62.3 & 13.4 \\
& Scale-MAE \cite{Reed_2023_ICCV} & V & fMoW & random & 92.9/93.9 & \textbf{70.8}/70.4 & 82.2/84.9 &  75.7/71.4 & 51.9 & 68.4 & 20.1 \\
& Scale-MAE & V & S2-LS & random & 83.4/89.5 & 64.3/71.7 & 70.2/79.8 & 67.5/66.3 & 32.3 & 61.5 & 13.1  \\
& Scale-MAE & V & S6 & random & 84.1/89.0 & 65.1/70.1 & 70.8/78.6 & 66.8/66.9 & 34.1 & 62.1 & 13.5 \\
& \textbf{X-STARS} & R & S2-LS & random & 77.3/94.6 & 57.6/69.4 & 72.2/89.1 & 54.4/70.3 & 36.1 & 62.1 & 18.5 \\
& \textbf{X-STARS} & V & S2-LS & random & 94.4/95.5 & 70.4/70.5 & 89.9/91.7 & 76.9/76.4 & 48.1 & 67.6 & 19.9 \\
\midrule
\multirow{5}{*}{\rotatebox[origin=c]{90}{Continual}} 
& {GFM} \cite{Mendieta_2023_ICCV} & S & GeoPile & ImageNet & 92.6/94.2 & 66.3/68.4 & 81.2/85.2 & 71.5/72.1 & 49.2 & 66.7 & 18.2\\
& \textbf{X-STARS} & V & LS$\rightarrow$S6 & X-STARS S2-LS& 95.0/\textbf{96.1} & 69.8/70.7 & 90.3/92.5 & 77.4/76.8 & 47.5 & 68.1 & 20.5 \\ 
& \textbf{X-STARS} & V & S2$\rightarrow$S6 & X-STARS S2-LS & \textbf{95.1}/95.3 & 69.8/69.5 & 91.9/91.9 & 78.2/76.6 & 45.3 & 68.1 & 19.7 \\ 
& \textbf{X-STARS} & V & S6$\rightarrow$S2 & Scale-MAE & 95.0/95.1 & \textbf{70.8}/69.7 & 91.1/92.4 & 77.8/77.2 & 46.5 & 68.7 & 19.8 \\
& \textbf{X-STARS} & V & S6$\rightarrow$BDORTHO & Scale-MAE & 91.9/94.3 & 69.4/\textbf{73.3} & \textbf{93.0}/\textbf{97.6} & \textbf{78.6}/\textbf{79.1} & \textbf{53.3} & \textbf{69.8} & \textbf{20.9} \\

\bottomrule
\end{tabular}
\end{adjustbox}
\end{table*}

We evaluate the representations learned by the models for various downstream tasks for land cover scene classification and land cover semantic segmentation. We freeze the backbone and perform either a non-parametric $k$ nearest-neighbor classification (k-NN), or a linear probing by training a linear classifier on top of the representations. This allows us to evaluate both whether semantically similar observations are grouped in the feature space, and whether the discriminative power of the representations is organized in a linear structure. For scene classification, we selected four datasets: EuroSAT \cite{helber2019eurosat}, Worldstrat \cite{cornebise2022open}, UC-Merced \cite{yang2010bag} and CV-BrCT \cite{machado2020airound}.
We report the top-1 accuracy on the downstream datasets using k-NN (averaged over $k=5, 10, 20, 50, 100, 200$) and linear probing.
For the semantic segmentation task, for the ViT, we fine-tune a UperNet segmentation head \cite{xiao2018unified} on top of our pretrained backbone. For the ResNet50, we fine-tune a decoder of a ResUNet, while freezing the pretrained encoder.
We selected three datasets: SpaceNet8 \cite{hansch2022spacenet}, Chesapeake Land Cover (CLC) for which we use Landsat-8 only \cite{robinson2019large} and the subset of FLAIR \cite{garioud2023flair} defined by \cite{Marsocci_2023_CVPR}. 
For this task, we report mIoU.
This choice of downstream tasks allows us to evaluate models on datasets with very different resolutions and sensor characteristics. For all datasets, we only consider the RGB bands.
In addition to the benchmark results, we perform ablations studies for the alignment loss $L_\text{MSAD}$ (\cref{sec:ab_msad} and \cref{sec:lambda}), impact of different input resolutions (\cref{sec:res}), combinations of input pairs (\cref{sec:combo}), impact of the knowledge distillation continual pretraining (\cref{sec:kd}), impact of model scale (\cref{sec:scale}) and few-shot experiments (\cref{sec:fs}). Experiments for these ablations are done with a ViT ``tiny'' backbone on 2 NVIDIA V100 GPUs and a batch size of 64/GPU.

\subsection{Comparison with state-of-the-art}
\label{sec:comp}

Table \ref{tab:comp} reports results for existing self-supervised EO models and our approach.

\paragraph{Training from scratch} For classification, we observe that X-STARS outperforms existing approaches on nearly all downstream tasks when trained on Sentinel-2 and Landsat-8, both with k-NN and linear probing. On average, X-STARS trained from scratch outperforms {both contrastive-based and MAE-based} models (\eg +1.3  \% on EuroSAT and +7.7\% on UC-Merced with respect to Scale-MAE). The multimodal alignment on low-resolution imagery (\ie Sentinel-2 and Landsat-8) learns effective representations for non-parametric k-NN and linear probing. {This is confirmed by the underperformance of vanilla DINO on MSC-France in average.} X-STARS outperforms Scale-MAE on three of the four classification tasks, Worldstrat being the exception. This could be expected since Scale-MAE is trained on the fMoW~\cite{christie2018functional} dataset, consisting mostly of WorldView and GeoEye imagery at $<$\SI{1}{\meter\per\pixel} GSD, which matches closely the \SI{1.5}{\meter\per\pixel} GSD from the SPOT-6 sensor used for Worldstrat. In comparison, the Sentinel-2 data used to train our model has a \SI{10}{\meter\per\pixel} GSD and X-STARS still manage a 70.4\% k-NN accuracy vs. 70.8\% for Scale-MAE. Note that this is despite MSC-France dataset containing only $\approx$5k patches, while Scale-MAE is trained on the 363k patches from fMoW~\cite{christie2018functional}, {CROMA on $\approx$255k multispectral patches from SSL4EO-S12 \cite{wang2023ssl4eo}} and SeCo on $\approx$200k images~\cite{manas2021seasonal}.

On segmentation tasks, X-STARS trained from scratch {either outperforms or is on-par with existing models}, only losing against Scale-MAE (-2.8\% mIoU on FLAIR, -0.8\% on SN-8, -0.2\% on CLC).
Scale-MAE obtains the best accuracy on segmentation across models trained from scratch, thanks to their effective scale-invariant strategy on high-resolution images, \eg mIoU of 51.9\% vs 48.1\% on the \SI{0.25}{\meter\per\pixel} aerial images from FLAIR. However,  the gap becomes smaller on lower resolution images, \ie SN-8 (\SI{0.8}{\meter\per\pixel} GSD) and CLC (\SI{30}{\meter\per\pixel} GSD).

Moreover, to show that the X-STARS improvements are not due to the newly proposed dataset, we also reported some experiments with {a vanilla DINO,} Scale-MAE and SatMAE\footnote{Without temporal or location metadata, \ie similar to a vanilla MAE.} trained from scratch on MSC-France (\ie Sentinel-2 and Landsat-8 pairs). For Scale-MAE we report also an experiment with only SPOT-6 images, to make it more similar to fMoW. As shown in \cref{tab:comp}, X-STARS outperforms these models. {The strong performance of the vanilla DINO, still inferior on average to X-STARS,} confirms that MIMs need either high-resolution or abundant data to work properly.%

\paragraph{Continual pretraining}
We also show that X-STARS can improve existing models using continual pretraining. Using the X-STARS S2-LS model, we adapt the model to SPOT-6 either using Landsat-8 or Sentinel-2 as a reference, improving the accuracy on most downstream tasks. However, we can also adapt an existing model, \eg using Scale-MAE as a domain teacher. Since Scale-MAE was trained on the very high-resolution images from the fMoW dataset, we use the SPOT-6 (\SI{1.5}{\meter\per\pixel}) as the ``source'' domain for the adaptation. 

Using Scale-MAE as initialization and adapting to Sentinel-2 images, X-STARS improves all the classification results (\eg for k-NN, average top-1 accuracy 94.6\% on EuroSAT and 70.8\% on Worldstrat), as low-resolution features are more effective for high-level semantics. Meanwhile, adapting to the aerial images from BDORTHO, we observe a significant boost in accuracy on segmentation datasets (\eg +1.4\% mIoU on FLAIR), and also the classification task on the high-resolution datasets (\eg +11\% on UC-Merced, +3\% on CV-BrCT). This demonstrates that X-STARS can be used to improve the performance of large-scale self-supervised EO models by adapting them to new sensors and finding representations better suited to downstream tasks.

{Finally, we compare X-STARS with the continual pretraining of GFM \cite{Mendieta_2023_ICCV}. X-STARS always outperforms GFM in linear probing and kNN. This is in part because GFM is designed for end-to-end fine-tuning, which is more expressive but also significantly more computationally expensive.}

\begin{table}
  \centering
  \caption{Accuracy on the validation test of downstream datasets with different configurations of the MSAD loss.}
  \label{tab:abl}
  \begin{tabular}[t]{cccccc}
  \toprule
  {MSAD} & {Patchwise} & {Label smoothing} & {EuroSAT} & {Worldstrat} & {SN-8} \\
  \midrule
         &            &                  & 91.6      & 72.5        & 63.6  \\
  \checkmark &        &                  & 91.9      & 72.0        & 64.1  \\
  \checkmark & \checkmark &            & 92.4      & 71.6        & 64.6  \\
  \checkmark &        & \checkmark      & 85.6      & 70.2        & 61.9  \\
  \checkmark & \checkmark & \checkmark  & \textbf{92.7} & \textbf{73.1} & \textbf{64.8} \\
  \bottomrule
  \end{tabular}
\end{table}

\subsection{Ablation on label smoothing and patchwise alignment}
\label{sec:ab_msad}

We report in Table \ref{tab:abl} downstream accuracies on X-STARS trained on Sentinel-2 + Landsat-8 in various configurations of the MSAD alignment loss. All the experiments were conducted on the validation sets of three datasets (EuroSAT, Worldstrat and SpaceNet-8) so as not to bias the results on the test sets.
First, we observe that using the MSAD alignment on the similarities computed on the global tokens already improves by $\approx0.6\%$ model performance both in classification (EuroSAT and Worldstrat) and segmentation (SpaceNet-8). 
 
Second, we observe an accuracy improvement brought by applying the alignment in a patchwise manner (PW). This shows the importance of learning to align tokens representing \emph{local regions} and not only the global image representation.
Third, we evaluate the impact of label smoothing (LS) in MSAD. Smoothing alone actually degrades performance compared to the baseline. However, when used in conjunction with patchwise alignment, smoothing improves performance.
This is because image pairs have the same \emph{global} semantics. Therefore, the smoothing adds noise to the learning process. However, semantics can be \emph{locally different}, especially with images at different resolutions. In that case, smoothing the similarities helps model softer alignments, resulting in better downstream performance. The patchwise mechanism is useful for learning local invariants, and the smoothing for grasping high-level similarities among different objects (\eg different forests or different crops).

\subsection{Impact of the $\lambda$ MSAD weight}
\label{sec:lambda}

\begin{figure}[t]
\includegraphics[width=.7\linewidth]{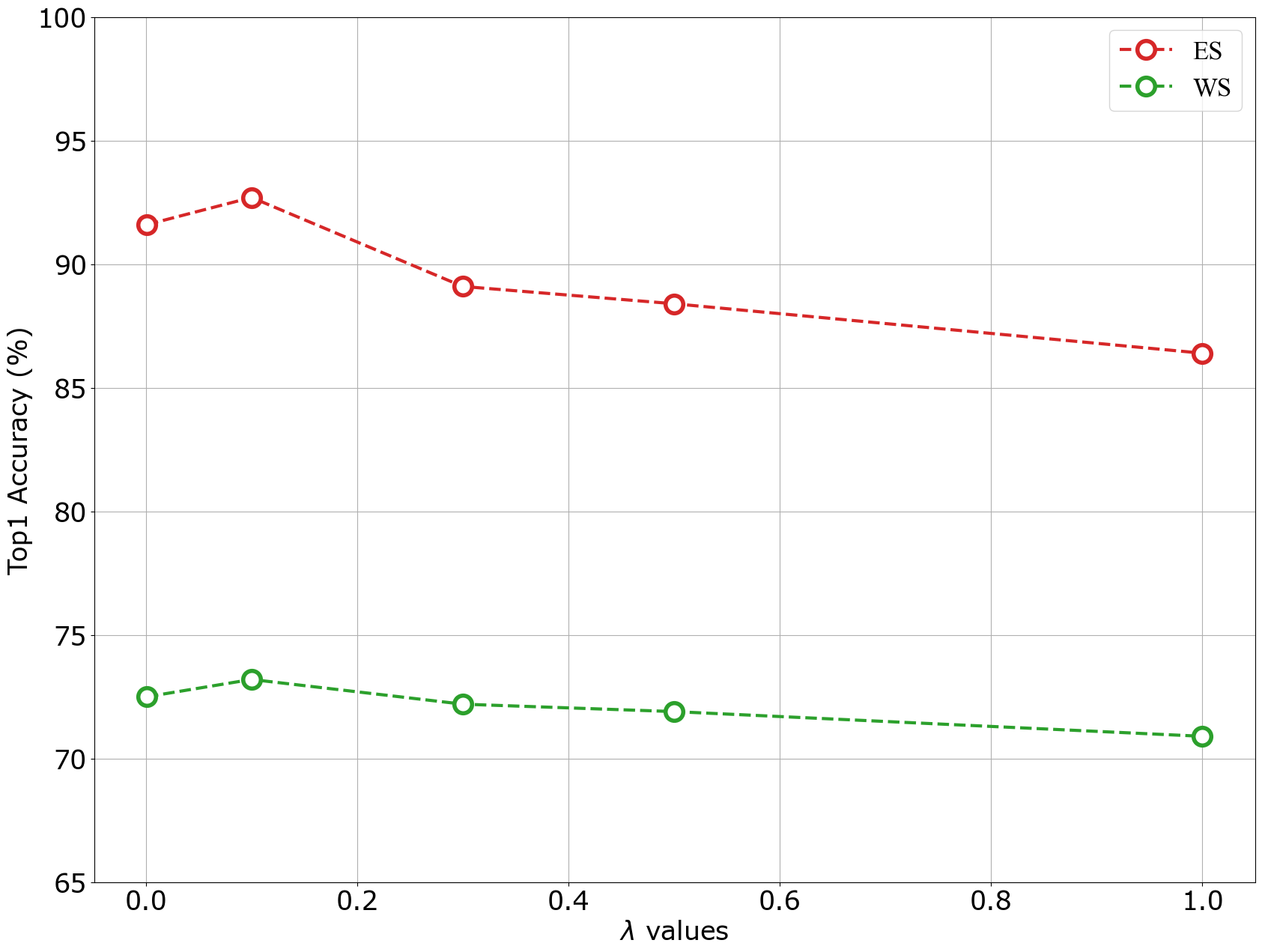}
\centering
\caption{Accuracy on downstream tasks (EuroSAT and Worldstrat) with various values for the $\lambda$ weight of MSAD loss.}
\label{fig:lambda}
\end{figure}

\cref{eq:xstars}, we introduced the $\lambda$ hyperparameter to control the relative importance of the MSAD alignment compared to the SSL objective. We report in \cref{fig:lambda} downstream accuracies for X-STARS with various values of $\lambda$. Note that $\lambda = 0$ is equivalent to a baseline model trained with DINO only. We observe that X-STARS outperforms the baseline for $\lambda \in [0, 0.3]$. The performance then decreases monotonically. The behavior is consistent for the two considered datasets (EuroSAT and Worldstrat). 
With high $\lambda$ values, the MSAD loss dominates the objective and the main SSL loss is not used anymore~\cite{caron2021emerging}, degrading performance and encouraging collapse across sensors. However, X-STARS is robust to the choice and $\lambda$ and $0.1$ resulted in improvements consistently over all the datasets.

\subsection{Impact of different input resolutions}
\label{sec:res}

The results in Table \ref{tab:res} assess the impact of resizing data on model performance, considering variations in resolution between pretraining and downstream tasks. In Table \ref{tab:comp}, the Sentinel-2 patches in MSC-France, at $300\times300$ with a GSD of \SI{10}{\meter\per\pixel}, were resized to $224\times224$ for pretraining, resulting in a resolution of $\approx$\SI{13}{\meter\per\pixel}. In contrast, the downstream EuroSAT task utilized $64\times64$ patches, equivalent to a GSD of $\approx$\SI{3}{\meter\per\pixel} with $224\times224$ patches.
Pretraining with a patch size the most similar to the original DINO yielded the best performance. This suggests looking for a trade-off between using the highest resolution and maintaining an appropriate patch size. Moreover, higher resolution generally leads to better performance, as shown also in \cite{corley2023revisiting}. The highest accuracy of 94.6\% is achieved when both pretraining and linear evaluation input sizes are $224\times224$ (Table \ref{tab:res}).
Interestingly, there appears to be no direct correlation between the GSD of pretraining and downstream tasks, despite using images from the same sensor.

\begin{table}
  \centering
  \caption{Impact of different resolutions (patch sizes) for S2 images in X-STARS pretraining on MSC-France.}
  \label{tab:res}
  \begin{tabular}[t]{ccc}
  \toprule
  {Training patch size} & {Evaluation patch size} & {EuroSAT accuracy} \\
  \midrule
  \multirow{2}{*}{224} & 224 & \textbf{94.6} \\
                        & 64  & 90.9          \\
  \midrule
  \multirow{2}{*}{100} & 224 & 93.7          \\
                        & 64  & 87.3          \\
  \midrule
  \multirow{2}{*}{300} & 224 & 92.9          \\
                        & 64  & 91.8          \\
  \bottomrule
  \end{tabular}
\end{table}

\begin{table}
  \centering
  \caption{Downstream accuracy after pretraining on different combinations of satellite images, with and without MSAD. 
  }
  \label{tab:combo}
    \begin{tabular}[t]{ccccc}
    \hline
    {Training sensors} & MSAD & {Epochs} & {EuroSAT} & {Worldstrat}   \\
    \midrule
    \multirow{3}{*}{LS-S2}                                     &           & 400    &  91.8       &   68.3             \\
    &            & 800    &    93.3     &    68.7               \\
    & \checkmark & 400    &   93.4  &    \textbf{70.8}      \\
    \midrule
    \multirow{3}{*}{S2-S6}                                     &           & 400    &      92.2   &     68.8            \\
    &            & 800    &     93.7    &    69.2               \\
    & \checkmark & 400    &   \textbf{93.8}      &    70.5      \\
    \midrule
    \multirow{3}{*}{LS-S6}                                     &           & 400    &      92.8   &    68.7               \\
    &            & 800    &   93.6      &      68.7             \\
    & \checkmark & 400    &    93.7     &      69.5          \\
    \midrule
    \multirow{3}{*}{\begin{tabular}[c]{@{}c@{}}S2-LS-\\ S6\end{tabular}} &     & 400    &   93.4      &    68.3     \\
    &            & 800    &   \textbf{93.8}      &   68.4       \\
    & \checkmark & 400    &    \textbf{93.8}     &    68.8      \\
    \hline
    \end{tabular}
\end{table}%

\subsection{Impact of training sensors}
\label{sec:combo}

To investigate the impact of the alignment MSAD loss, we evaluate several models trained on different sensor combinations. Results are reported in Table \ref{tab:combo}. Without the MSAD, the model is trained alternatively with batches of one sensor or the other, with no specific alignment. We observe that, with MSAD, the model i) converges faster; ii) reaches higher downstream accuracy on average. However, the current implementation of X-STARS saturates when three or more sensors are trained together. Indeed, training on the S2-LS-S6 combination does not perform better than any other combination of the two. We hypothesize that this is due to the pairwise computation of the MSAD alignment. Indeed, in this setup, we compute all pairwise similarity pairs (\ie S6-S2, LS-S2, S2-S6), which are then averaged over. This could allow for inconsistent similarities between pairs of sensors, slowing down the convergence overall. Nonetheless, multimodal alignment with MSAD consistently improves the baseline in all two sensor combinations.

\subsection{Different continual pretraining data strategies}
\label{sec:kd}

In this ablation, we evaluate the improvements brought by the ``continual pretraining'' strategy. We adapt an existing pretrained model using the MSAD loss. Models are evaluated on two downstream tasks (EuroSAT and Worldstrat). Results are reported in Table \ref{tab:kd}, where $A\rightarrow B$ means that the model is pretrained for 400 epochs on sensor $A$ and then adapted for 400 epochs on sensor $B$. We also evaluate the $A\rightarrow A$ setting, which is standard ``same sensor'' knowledge distillation.
Knowledge distillation in itself improves significantly the downstream model performance, even with the same sensor. However, in all cases, cross-sensor alignment improves the model even further. 
Generally speaking, when employing higher resolution images we can obtain slightly better results (\eg 69.5\% average top-1 accuracy for Worldstrat and 93.6\% for EuroSAT). While adapting tends to improve performance on average, the gains are the most significant when the resolution of the downstream dataset matches the resolution of the images used for the adaptation. On average, EuroSAT results are improved by adapting on Sentinel-2, Worldstrat results are improved by adapting on SPOT-6, FLAIR results are improved by adapting on BDORTHO, etc. In conclusion, there is a trade-off between mixing features from different sensors and generality, that can be dealt with by adapting on the appropriate sensor.

\begin{table}
  \centering
  \caption{Downstream accuracy after different combinations of adaptation through knowledge distillation.}
    \label{tab:kd}
  \begin{tabular}[t]{cccc}
  \toprule
{Pretraining} & Adaptation & {EuroSAT} & {Worldstrat}       \\
\midrule
LS        & $\varnothing$      &  86.0   &   64.2            \\
LS        & $\rightarrow$ LS &   91.5  &    68.0          \\
LS        & $\rightarrow$ S2 &  92.6   &  69.1             \\
LS        & $\rightarrow$ S6 &  92.3   &    69.3           \\
\midrule
S2       & $\varnothing$      &  86.2   &  64.1             \\
S2       & $\rightarrow$ S2 &   92.2  &   68.2            \\
S2       & $\rightarrow$ LS &  93.2   &  68.4             \\
S2       & $\rightarrow$ S6 &  92.7   &    69.4           \\
\midrule
S6           & $\varnothing$        & 82.2    &  63.5            \\
S6           & $\rightarrow$ S6 &   91.7  &    68.4           \\
S6           & $\rightarrow$ LS &   92.8  & \textbf{69.5}       \\
S6           & $\rightarrow$ S2 &    \textbf{93.6} &   69.1      \\
\bottomrule
  \end{tabular}
\end{table}

\begin{table}[h!]
\centering
\caption{Results for X-STARS, trained in a self-supervised way from scratch with different ViT backbone. {Each column name denotes one dataset, similarly to Table \ref{tab:comp}.}}
\begin{adjustbox}{width=0.5\textwidth}
\label{tab:scale}
\setlength\tabcolsep{3pt}
\begin{tabular}{cccccccccc}
\toprule
Backbone & ES & WS & UC-M & CV-BrCT & FLAIR & SN8 & {CLC} \\ \midrule
ViT-T/16 & 93.4 & \textbf{70.5} &  85.9 &  73.7 & 41.1 & 64.6 & 18.7 \\
ViT-B/16 &  \textbf{95.2} &  70.2 &  89.4 &  76.2 & 42.6 & 65.8 & 19.1 \\
ViT-L/16 & 94.4 & 70.4 & \textbf{89.9} & \textbf{76.9} & \textbf{48.4} & \textbf{67.6} & \textbf{19.9}  \\
\bottomrule
\end{tabular}
\end{adjustbox}
\label{tab:mod_size}
\end{table}

\subsection{Model scale}
\label{sec:scale}

To assess the effectiveness of X-STARS, we also tried to vary the size of the backbone, for the self-supervised training from scratch framework. The results are presented in Table \ref{tab:mod_size}. Among the results, we can see that in average, bigger transformers lead to better results. However, there are some interesting phenomena. Increasing the depth of the transformer (tiny/base vs. large) improves the performance in segmentation more clearly (ViT-L/16 gains +5.8\% mIoU on FLAIR). On the other side, for classification tasks, we observe a clear correlation with dataset resolution. For datasets with lower resolution, even lighter transformers perform very robustly. In general, we can assert that our X-STARS reaches performance comparable with state-of-the-art results, also with a small model.

\begin{table}[]
\centering
\caption{Few-shot learning results on EuroSAT. The reported results are Top-1 accuracy (\%)}
\label{tab:fs}
\begin{tabularx}{0.4\textwidth}{lYYY}
\toprule
Method/Data \% & 5\% & 10\% & 50\%\\ 
\midrule
Scale-MAE & 75.2 & 79.7& 86.7\\
X-STARS (pretraining) & 88.1 & 91.4& \textbf{94.6}\\
X-STARS (continual) & \textbf{89.6} & \textbf{91.5}& 94.5\\
\bottomrule
\end{tabularx}
\end{table}

\subsection{Low-data regime experiments}
\label{sec:fs}

To better assess the effectiveness of X-STARS, we also investigated the few-shot capabilities of our model. Specifically, we selected the best competitive model, \ie Scale-MAE, and the best X-STARS under the two different approaches (\ie continual and pretraining), and performed linear evaluation with an increasing number of labeled data available (\ie 5\%, 10\%, 50\%) for EuroSAT. The results, shown in Table \ref{tab:fs}, show the superiority of X-STARS.

\section{Conclusion}
\label{sec:conc}

We introduced X-STARS, a cross-sensor alignment framework for RS. We designed MSAD, a dense contrastive loss that performs cross-sensor knowledge distillation to learn \emph{sensor agnostic} representations of EO images, that also take into account \emph{local semantics}. X-STARS achieves on-par results with existing self-supervised models such as ScaleMAE while using 10 to 100$\times$ less data. Moreover, X-STARS objective can be used for continual pretraining to improve existing models by adapting them to sensors that are more suited to the targeted downstream tasks. We establish new SOTA for SSL on multiple downstream classification and segmentation tasks. Ablation studies demonstrate the benefits of the patchwise alignment and the multimodal loss, especially on segmentation tasks which are more challenging for self-supervised models than classification.
X-STARS can be extended to integrate more challenging modalities, that so far weren't considered for their inherent greater complexity \eg multispectral, SAR, thermal or even vector geodata.
{Another interesting research line interests how to adapt effectively the ViT patch embedder to deal with different spatial resolutions. Some first attempts were made in this direction \cite{beyer2023flexivit, han2024bridgingremotesensorsmultisensor}, but the problem still remains open.}
While the adaptation through continual pretraining is effective, efficiency could be improved by using better model adaptation such as low-rank adaptation as popularized for large language models~\cite{hu2021lora}. This would allow for seamless transitions of representations from one sensor to another without the need for end-to-end fine-tuning. 

\section*{Acknowledgements}
The authors thank Devis Tuia for fruitful discussions on the design and scaling of self-supervised losses. This project was made possible thanks to the financial support of the ANR MAGE (ANR-22-CE23-0010) and Google for their donation under the Research Scholar program.
This work was granted access to the HPC resources of IDRIS under the allocation AD011014518 made by GENCI. 

\bibliographystyle{ieeetr}
\bibliography{IEEEexample}

\clearpage

\appendices
\section{Downstream Dataset Details}

In Table \ref{tab:data}, we reported the main characteristics of the datasets used for the downstream tasks. We can note a high diversity in the number of classes, patch size, GSD and number of images. This shows the potentiality of X-STARS.
For EuroSAT, that is made of Sentinel-2 MSIs, we used just the RGB images, and used the split already prepared in \cite{helber2019eurosat}.
For Worldstrat, shaped originally for super-resolution, we selected only the Airbus SPOT6/7 images. We used the eight LCCS (land cover classes), provided in the metadata, as scene classification labels. Also, the stratified split is offered by the authors \cite{cornebise2022open}. We used this dataset, because we needed a downstream task on SPOT images, to validate our ideas.
For UC-Merced, we used the images as they are. We performed a custom split, given the absence of an official one and the variability of the usage.
For CV-BrCT, we selected the subset with the aerial images. We discarded the ground images. We selected this dataset, due to its high variance in the resolution of the aerial images. We performed a custom split, for the same reasons as for UC-Merced.
Concerning the segmentation datasets, for FLAIR, we followed the conventions utilized in \cite{Marsocci_2023_CVPR}. We took a subset of 13 domains (10 for training and 3 for testing), and we trained the models on $256\times256$ patches.
For SpaceNet8, we selected only the pre-flood images insisting on Louisana area. We used $256\times256$ patches and we adopted the split presented in \cite{hansch2022spacenet}.
For Chesapeake Land Use dataset, we needed a dataset made just of Landsat images. For this reason, among all the data available, we selected just the low-resolution Landsat data, consisting in $200\times200$ patches. The high-resolution were no used at all. This means that the dataset has really poor information, as shown by the average of the metrics ($\sim20\%$ mIoU). We adopted the original split.

\begin{table}[h!]
\centering
\adjustbox{height=1.1cm}
{
\begin{tabular}{cccccc}
Dataset     & Patch Size (px) & GSD (m) & N. Classes & N. Images & Task \\
    \hline
EuroSAT  \cite{helber2019eurosat}   & 64 & 10 & 10 & 27,000 & C \\
Worldstrat \cite{cornebise2022open} & 1054 & 1.5 & 7 & 3,823 & C \\
UC-Merced  \cite{yang2010bag} & 256 & 0.3 & 21 & 2,100 & C \\
CV-BrCT   \cite{machado2020airound}  & 500 & 0.3-4800 & 9 & 24,000 & C \\
FLAIR   \cite{garioud2023flair}    & 256 & 0.2 & 14 & 21,396 & S \\
SpaceNet8 \cite{hansch2022spacenet}  & 256 & 0.5 & 3 & 14,975 & S \\
Chesapeake \cite{robinson2019large} & 200 & 30 & 15 & 731 & S \\
\end{tabular}
}
\caption{Dowsntream datasets' info. C stands for Classification, S for Semantic Segmentation}
\label{tab:data}
\end{table}

\section{Downstream training process details}
\label{sec:train}

Both for linear probing and semantic segmentation, we used a single NVIDIA RTX 6000 GPU. 
For linear probing, we trained the linear layer for 100 epochs, with a learning rate of 0.01. We used a batch size of 32.

For semantic segmentation, we trained an UperNet decoder, freezing the backbone, for 100 epochs. An early stopping stops the training after 25 epochs of patience. We fixed 32 as the batch size and 0.01 as the learning rate. 

\begin{figure}[ht]
\centering
\includegraphics[width=.9\linewidth]{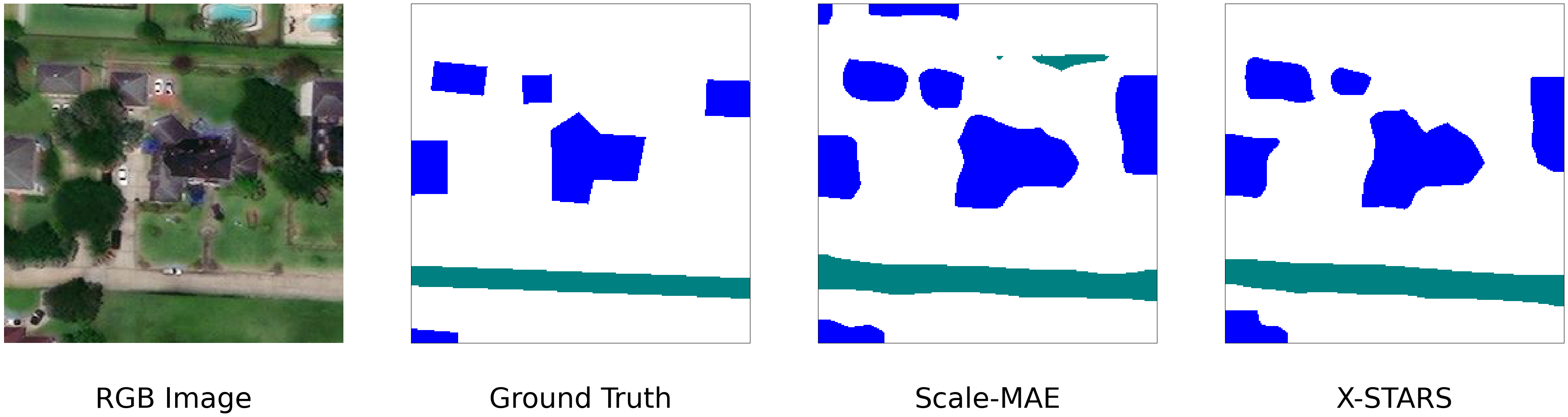}
\includegraphics[width=.9\linewidth]{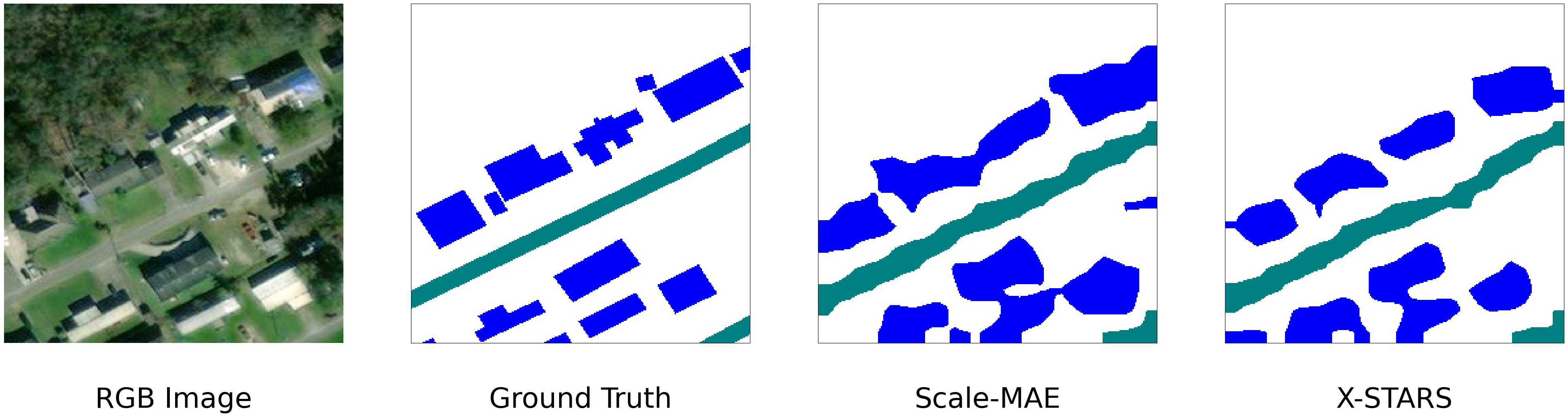}
\caption{Few examples from SpaceNet8 predictions.}
\label{fig:sem_segs}
\end{figure}

\begin{figure}[ht]
\centering
\includegraphics[width=.9\linewidth]{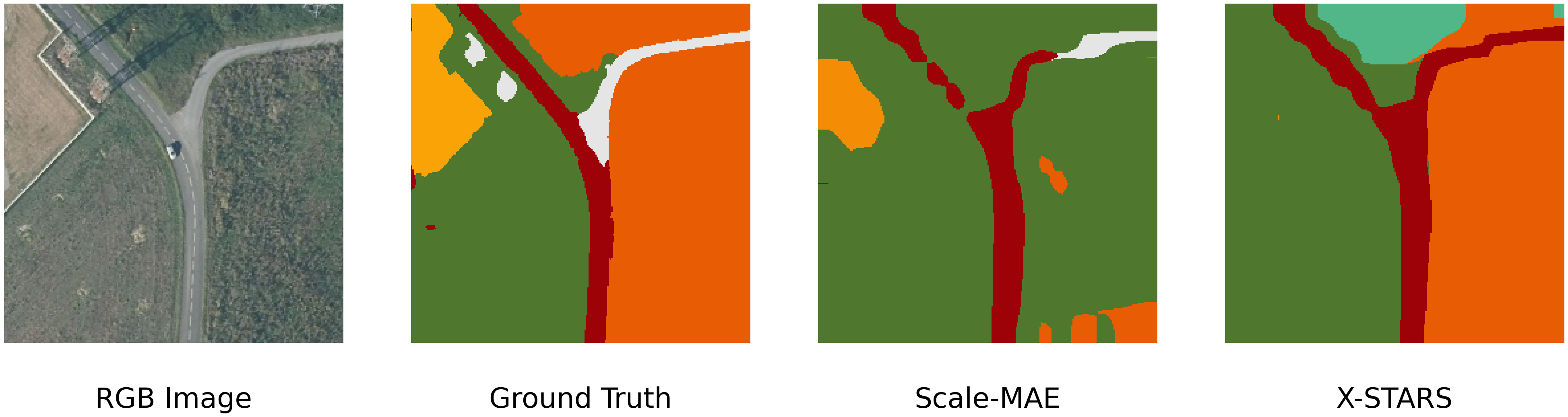}
\includegraphics[width=.9\linewidth]{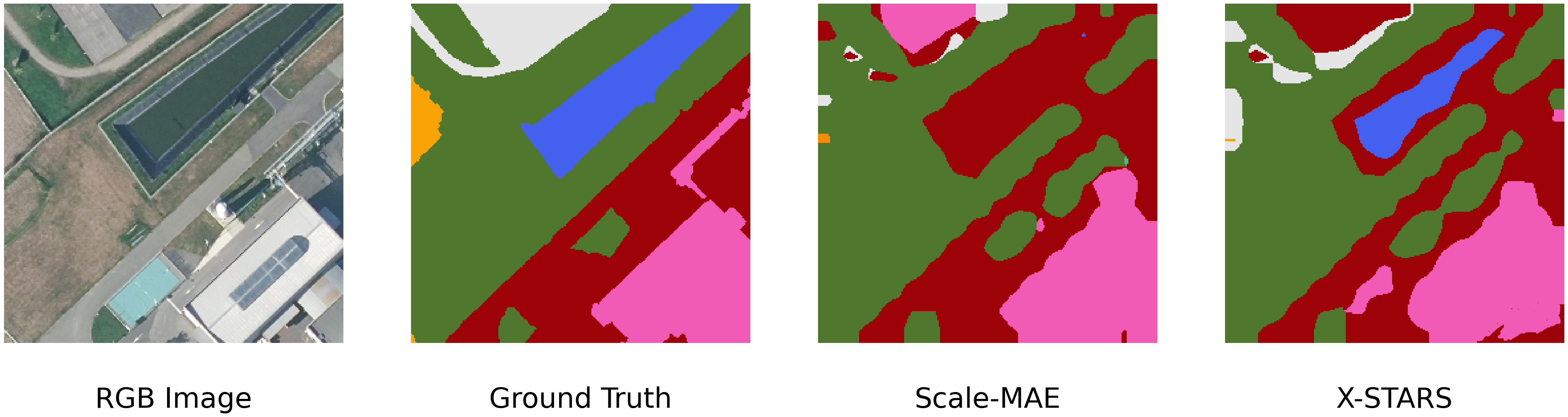}
\caption{Few examples from FLAIR predictions.}
\label{fig:sem_segf}
\end{figure}

Finally, for segmentation, we show some results for SpaceNet8 (Figure \ref{fig:sem_segs})and FLAIR (Figure \ref{fig:sem_segf}), to show the robustness of the adapted features of X-STARS, w.r.t. Scale-MAE.

\subsection{Finetuning}
\label{sec:ft}

Table \ref{tab:ft} presents the fine-tuning results on the EuroSAT dataset, with accuracies reported from \cite{cong2022satmae} and \cite{wanyan2023dinomc}. The supervised baseline trained on ImageNet (Sup. IN) achieves an accuracy of 86.4\%, providing a lower-bound for comparison. Self-supervised approaches, such as SeCo \cite{manas2021seasonal} and SatMAE \cite{cong2022satmae}, yield significant improvements, with accuracies of 93.1\% and 95.7\%, respectively. DINO-MC, which leverages a vision transformer with multi-crop training, further enhances performance, reaching 98.8\%. X-STARS, trained for 100 epochs using AdamW with a learning rate of 0.0001, achieves an accuracy of 99.1\%, outperforming all other methods. This result demonstrates the superior capability of X-STARS in fine-tuning for satellite image classification. The consistent improvement over existing methods highlights the effectiveness of X-STARS in leveraging both self-supervised learning and domain-specific adaptations.

\begin{table}[]
\caption{Results, expressed in terms of accuracy (\%), of finetuning on EuroSAT. Results are reported from \cite{cong2022satmae} and \cite{wanyan2023dinomc}.}
\label{tab:ft}
\centering
\begin{tabularx}{0.4\textwidth}{lYYYYY}
\toprule
Dataset/Method & Sup. (IN) & SeCo & SatMAE & DINO-MC & X-STARS \\ \midrule
EuroSAT     & 86.4      & 93.1 & 95.7   & 98.8    & \textbf{99.1}    \\ \bottomrule
\end{tabularx}
\end{table}

\ifCLASSOPTIONcaptionsoff
  \newpage
\fi

\end{document}